# Team Plan Recognition: A Review of the State of the Art

**Loren Rieffer-Champlin**

University of Arizona, Tucson, AZ 85721, USA

**ABSTRACT**

There is an increasing need to develop artificial intelligence systems that assist groups of humans working on coordinated tasks. These systems must recognize and understand the plans and relationships between actions for a team of humans working toward a common objective. This article reviews the literature on team plan recognition and surveys the most recent logic-based approaches for implementing it. First, we provide some background knowledge, including a general definition of plan recognition in a team setting and a discussion of implementation challenges. Next, we explain our reasoning for focusing on logic-based methods. Finally, we survey recent approaches from two primary classes of logic-based methods (plan library-based and domain theory-based). We aim to bring more attention to this sparse but vital topic and inspire new directions for implementing team plan recognition.

## INTRODUCTION

Most approaches presented in the plan recognition literature focus on inferences about the plans of a single human agent (Zhuo, 2019). However, there is a need for artificial intelligence (AI) technology to implement plan recognition for a coordinated group of agents with a shared objective (Falcone and Castelfranchi, 1996; Zhuo, 2019). We refer to this specific case as Team Plan Recognition (TPR). Despite the usefulness of TPR, the literature could be more extensive. Therefore, we aim to draw more attention and interest to this vital subject by surveying recent logic-based methods for TPR.

More specifically, we examine plan recognition literature that explicitly proposes logic-based implementations in team settings where each team member is modeled as an individual agent. Thus, excluding implementations not designed with multiple agents in mind or ones that treat a group as a single agent. Additionally, we prefer to use the term TPR instead of multi-agent plan recognition, contrary to the literature we are reviewing. We do this to emphasize that our study focuses on implementing plan recognition on agents working together and coordinating as opposed to those that are non-coordinating or even adversarial towards each other.

We organize the paper as follows: the next section defines TPR in the context of the general plan recognition field of research and summarizes implementation challenges. The third explains our reasoning for focusing on logic-based methods. The fourth and fifth sections survey approaches from two distinctive classes of logic-based methods, plan-library based and domain-theory based, respectively. Finally, the last section concludes this review and gives further research directions.

## WHAT IS A TEAM PLAN RECOGNITION SYSTEM?

A TPR system attempts to infer the latent structures and dependencies within a team's behavior given an observed sequence of actions (i.e., a plan) or other related information (e.g., state changes) (Falcone and Castelfranchi, 1996; Banerjee, Kraemer, and Lyle, 2010; Zhuo, 2019). The output of such a system is known as a plan explanation or hypothesis, which can take many forms depending on the system's implementation (Banerjee, Kraemer, and Lyle, 2010). Figure 1 illustrates doing TPR for a firefighting operation.

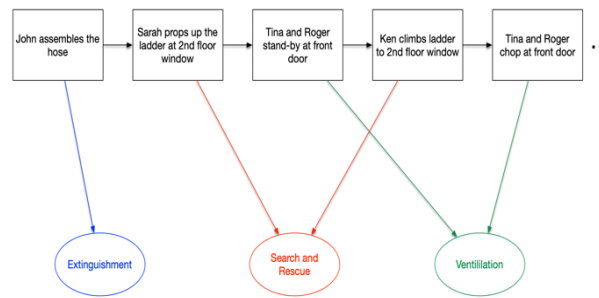

**Figure 1:** Example of TPR system inferring how observed actions (top) relate to a firefighting operation's goals (bottom).

One significant challenge of implementing TPR is observing and modeling multiple agents simultaneously. Primarily, this means handling an increased volume of observations and coinciding actions (Falcone and Castelfranchi, 1996; Banerjee, Kraemer, and Lyle, 2010; Banerjee, Lyle, and Kraemer, 2015). The latter can cause the space of possible plans to increase dramatically, and tracking them requires unique temporal reasoning mechanisms. Thus, TPR usually has a much higher computational cost than single-agent plan recognition.

Several unique types of inferences can also be made in a team setting. For example, a recognizer can do sub-team detection, which attempts to identify when and how teams might partition themselves into sub-teams (Banerjee, Kraemer, and Lyle, 2010; Sukthankar and Sycara, 2011). These sub-teams can then be linked to specific sub-goals, as also suggested in figure 1.

## WHY FOCUS ON LOGIC-BASED METHODS?

A previous literature review on activity, plan, and goal recognition methods by Van-Horenbeke and Peer (2021) presented four categories of approaches. Logic-based, classical machine learning, deep learning, and brain-inspired methods. Of these categories, 17 logic-based plan recognition implementations were surveyed out of 21. A similar phenomenon occurs in older reviews, demonstrating that logic-based methods are the most prevalent (Carberry, 2001; Armentano and Amandi, 2007). The cause is likely to be both historical precedents and viability.

A group of psychologists and AI researchers originally conceived the plan recognition problem when they attempted to use tools and methods developed in the AI field to test theories on "how human observers understand the actions of others" (Schmidt, Sridharan, and Goodson, 1978). At the time, these techniques were primarily built upon symbolic logic and reasoning. Consequently, plan recognition has been continually studied using mainly logic-based methodologies.

Nonetheless, there have been several attempts to use non-logic-based techniques, in particular, machine learning and deep learning. However, non-logic-based approaches are rarely as successful and are usually far less versatile than logic-based approaches, only able to handle straightforward and specific problem scenarios (Van-Horenbeke and Peer, 2021). This is because machine learning and deep learning methods are generally incapable of representing highly complex logical structures. Moreover, they typically require a vast amount of data when the logic-based methods need very little or none, at least in the traditional sense.

This viability likely extends to all variants of plan recognition, including TPR. Likewise, we also found a preference for logic-based TPR methods in our preliminary research. Accordingly, we narrowed our review to these methods specifically.

## PLAN-LIBRARY BASED APPROACHES

### Different Forms of Plan Libraries

As the name suggests, plan-library based implementations use a pre-constructed data structure called a plan library (Carberry, 2001; Armentano and Amandi, 2007; Van-Horenbeke and Peer, 2021). The form of this library can range from a collection of plan constructs to something more abstract and compact (Vilain, 1990). For example, Banerjee, Kraemer, and Lyle (2010) and Zhuo (2019) have defined their plan libraries as collections of matrices, each representing a non-specific sub-team performing a set of actions over a non-specific sequence of discrete time. For Banerjee, Kraemer, and Lyle (2010), the actions are grounded. Alternatively, Zhuo (2019) uses ungrounded actions and allows for missing actions. In either case, these matrices can represent sub-goals and the resource requirements (i.e., the number of agents and discrete time steps) for those sub-goals.

Another example would be the collection of AND/OR trees used by Sukthankar and Sycara's (2011) TPR implementation. They interpret each tree as representing a plan that requires a specified number of agents to execute. Furthermore, the trees have other particular nodes besides the AND, OR, and action nodes (called team behaviors). These nodes represent the redistribution of agents among sub-teams. In particular, SPLIT nodes specify that the team needs to split into sub-teams to execute smaller sub-plans as part of the plan containing the SPLIT. These sub-plans are other AND/OR trees matching the resource requirements of the SPLIT. Likewise, the plan with the SPLIT node may itself be a sub-plan to another plan. The concept is similar to the hierarchical task network (HTN) planning paradigm, where complex tasks are decomposed into sets of more straightforward tasks (Russell and Norvig, 2021).

### Problem Specifications and Methodologies

Most plan-library based approaches specify their problem as a particular case of the exact cover (EC) problem (Banerjee, Kraemer, and Lyle, 2010; Banerjee, Lyle, and Kraemer, 2015). We can define the general problem as follows: Given a plan library $\mathcal{L}$ and observation trace T, find a subcollection $\mathcal{L}'$ of $\mathcal{L}$ and partition P of T such that the members of P have a one-to-one correspondence with the members of $\mathcal{L}'$. The subcollection $\mathcal{L}'$ is then said to be the plan explanation for T. Lastly, all plan-library based approaches we surveyed are meant to be used offline, so we can assume that T covers a team's entire activity from start to finish.

For Banerjee, Kraemer, and Lyle (2010), T is an observation matrix of grounded actions where the columns represent individual agents, and the rows are discrete time steps starting at time 0. $\mathcal{L}$ is a collection of matrices, as mentioned in the previous section. Given the connection between TPR and EC, they find an exact solution by modifying a known approach to EC, Knuth's (2000) algorithm X. In addition to this, they created a branch and bound version of it to increase efficiency.

Banerjee, Lyle, and Kraemer (2015) prove that other plan libraries could be transformed into a collection of grounded action matrices, allowing them to solve many TPR problems using Banerjee, Kraemer, and Lyle's (2010) modified algorithm X. However, they also present a parsing-based algorithm for when $\mathcal{L}$ is a context-free grammar. Their algorithm adapts Vilain's (1990) implementation for single-agent plan recognition.

Zhuo (2019) defines T similarly, but like their $\mathcal{L}$, they allow missing observations in T. Moreover, any subcollection $\mathcal{L}'$ of $\mathcal{L}$ (i.e., plan explanations candidates) must be grounded according to a given domain description $\mathcal{A}$, initial state $s_0$, and set of goal states $G$. Likewise, all three of these can also be "incomplete," missing various pieces of information (e.g., missing predicates in action effects).

Furthermore, Zhuo (2019) initially framed their problem as a sizeable Boolean satisfiability (SAT) problem. However, recognizing that incomplete and missing information can make this problem infeasible, they relax it to a max satisfiability (MAX-SAT) problem. Then, appropriately, they estimate the best solution using Li et al.'s (2009) MAX-SAT solver. Consequently, Zhuo's (2019) approach is much more versatile (i.e., it can handle more types of problem scenarios) than Banerjee, Kraemer, and Lyle's (2010) but is susceptible to inference error.

Sukthankar and Sycara's (2011) plan recognition approach focused on a pre-processing technique to prune their AND/OR tree-based library. They assume that T is a set of traces, each specifying a sequence of team behaviors (e.g., advance, defend, attack, etc.) and the team of agents performing this sequence. Unlike the other approaches discussed so far, T is not observed. Instead, it is inferred from spatio-temporal data using a method for activity recognition also proposed by Sukthankar and Sycara (2011). The details of this method are outside the scope of this review, so we will leave it to the reader to follow up.

As implied by Sukthankar and Sycara's (2011) definition of $\mathcal{L}$, a trace $t_b \in$ T could contain a sub-plan and sub-team of a trace $t_a \in$ T. Therefore there is an implicit temporal dependency between these two traces. Likewise, there are dependencies between observations within individual traces. Their approach exploits these dependencies to remove trees from their library, thus, creating a smaller search space for their search algorithm. Finally, like Zhuo's (2019) approach, Sukthankar and Sycara's (2011) can handle noisy or missing observations.

## Summary

Despite similar problem formulations, most approaches have a unique methodology. For example, some approaches, like Sukthankar and Sycara's (2011), focus on pruning the library, whereas others concentrate on efficient search algorithms. Furthermore, plan-library based methods tend to rely on simplistic knowledge representations that lack expressive power, such as propositional logic, severely limiting the versatility of problem scenarios they can handle. More versatility, however, can be achieved by incorporating a domain description, first-order logic, and logical constraints, as noticed with Zhuo's (2019) approach. As discussed in the next section, these are some of the core elements of the domain-theory based class of methods. However, it is still being determined how these elements affect the efficiency of Zhuo's (2019) approach compared to other plan-library based approaches.

Lastly, Banerjee, Lyle, and Kraemer (2015) had the only recent work that featured a compact grammar and parsing-based algorithm. This is a surprising result given that much of the field, especially single-agent plan recognition, has been built upon using compact grammars and parsing techniques. For this reason, plan-library based methods are often referred to as *plan recognition as parsing* (Vilain, 1990; Van-Horenbeke and Peer, 2021). Despite this, this methodology is less prevalent in recent TPR literature.

## DOMAIN-THEORY BASED APPROACHES

### Planners and Knowledge Representation

Domain-theory based approaches are generative methods that attempt to reconstruct the observed plan using an automated planner. For this reason, these methods are also referred to as *plan recognition as planning* (Ramírez and Geffner, 2009; Van-Horenbeke and Peer, 2021). Upon successfully rebuilding the observed plan, the constructs and parameters used by the planner are returned as feasible plan explanations. As suggested, this class of methods requires domain and problem descriptions to be engineered.

Most of these approaches apply off-the-shelf planners (i.e., planning algorithms developed by someone else) rather than creating new algorithms (Ramírez and Geffner, 2010). For instance, Argenta and Doyle's (2017) method can use any planner, but they prefer Pellier's (2014) java implementation of GraphPlan. This implementation takes domain and problem descriptions written in PDDL as input.

Shvo, Sohrabi, and McIlraith (2018) present three different methods for TPR. These methods require a classical temporal planner and a diverse temporal planner. The latter finds a set of plans within some distance of each other using a given metric. For testing, they distinctly used Gerevini et al.'s (2014) LPG-TD and Nguyen et al.'s (2012) LPG-d, respectively. Accordingly, they likely use PDDL, the preferred action description language for the original implementations of LPG-TD and LPG-d.

Shvo et al. (2020) use epistemic planning in their approach. This reasonably recent planning paradigm attempts to model the agents' belief systems and knowledge acquisition. To illustrate, an agent needing to deliver a package must acquire knowledge of the drop-off address. An epistemic planner, in this case, must devise a plan that has the agent acquiring this knowledge to complete their goal. Furthermore, the knowledge must be achieved using a method that aligns with the agent's beliefs. This is to avoid mistakes, such as the agent obtaining a false address or disregarding the needed address.

Shvo et al.'s (2020) approach was tested using three different epistemic planners: Muise et al.'s (2015) RP-MEP, Huang et al.'s (2017) MEPK, and Le et al.'s (2018) EFP. As input into these planners, they specify their problem scenario using a multi-agent modal logic called $KD45_n$. As suggested, this logic has structures that can represent an agent's beliefs.

## Problem Specifications and Methodologies

Shvo, Sohrabi, and McIlraith (2018) and Shvo et al. (2020) define a plan explanation as the pair $(\pi, G)$, where $G \in \mathcal{G}$ is a goal in the set of achievable goals. $\pi$ is a plan that satisfies a sequence of observations $O$. Furthermore, $\pi$ must be valid under some domain description and achieve at least one goal in $\mathcal{G}$. Although it is not ideal, this goal can differ from $G$. This scenario is examined further by Shvo et al. (2020) in the context of differing beliefs between the observer and agents.

As with other approaches, the objective is to find the most likely plan explanation. For this purpose, Shvo, Sohrabi, and McIlraith (2018) proposed three methods for computing the posterior distribution, $P(G|O)$. These methods referred to as delta, diverse, and hybrid, rely on different types of planners (as suggested in the previous section). In particular, Shvo et al. (2020) use the delta method with their epistemic planner. The exact details of these methods will be left up to the reader to explore. Finally, once $P(G|O)$ is calculated (using any of the three methods), it can be used to find the team's most likely goal given their observations by computing the maximum a posteriori (MAP) estimate, $G'$.

As implied, $O$ is not necessarily a fully observed plan (i.e., sequence of actions). For example, Shvo, Sohrabi, and McIlraith (2018) assume that only the state changes (also called fluent) from the team's activity are observed. Moreover, $O$ may be missing observations related to the team's actions or might contain unexplainable observations (i.e., observations not caused by the team's actions). For this reason, each observation in $O$ must either be explained by an action or discarded as unexplainable. Thus, the most likely plan $\pi'$ that satisfies $O$ explains nearly all the observations, minimizing removed actions. At the same time, $\pi'$ should only include enough extra actions (i.e., actions that do not match observations in $O$) to make it valid. Such a $\pi'$ optimally satisfies $O$. However, $\pi'$ achieving $G'$ (as previously defined) is also ideal. Accordingly, these approaches favor a plan explanation $(\pi', G')$, where $\pi'$ optimally satisfies $O$ as much as possible given that it achieves $G'$.

Lastly, Argenta and Doyle (2017) propose an online TPR method called Probabilistic Multi-Agent Plan Recognition as Planning (P-MAPRAP). As suggested, P-MAPRAP can produce a likely plan explanation based on a partial sequence of observations starting from time 0 up to the current time. The observations are grounded actions parameterized by a specific agent.

They define a plan explanation as an interpretation comprised of partial interpretations. Each partial interpretation is a unique assignment of agents to a goal. An example given by Argenta and Doyle (2017) is $(A_Y, A_Z : G_1)$ which has agents $A_Y$ and $A_Z$ pursuing a goal $G_1$. Similar to Sukthankar and Sycara's (2011) methodology, Argenta and Doyle's (2017) interpretations attempt to partition the team's plan into sub-plans and the team into sub-teams. However, in this case, sub-plans are generated ad-hoc by a planner rather than retrieved from a plan library, and the team's entire plan is not yet observed. Therefore, the sub-plans are generated according to the partial interpretations and collectively must satisfy the observations seen so far. For instance, $(A_Y, A_Z : G_1)$, would have agents $A_Y$ and $A_Z$ executing a sub-plan that achieves $G_1$ and contains any observations involving these two agents.

Argenta and Doyle (2017) assume that the likelihood of an interpretation is inversely proportional to the accumulated costs of the sub-plans that satisfy it. These sub-plans are optimized as much as possible under the constraint that they collectively satisfy the current observations. For that reason, new sub-plans should be generated with each new observation to continually meet this constraint, causing the likelihood of interpretations to change over time. However, rather than recomputing the likelihood of all possible interpretations for each new observation, they increased the efficiency of their method by implementing a priority queue system that re-evaluates interpretations selectively. We will leave the exact details of this system to the reader.

## Summary

One of the nuances of domain-theory based methods is that their versatility relies heavily upon the expressivity of the knowledge representation used to model the problem scenario (i.e., the type of planning used) (Armentano and Amandi, 2007; Van-Horenbeke and Peer, 2021). Hence, Shvo et al. (2020) present one of the most versatile approaches we surveyed. The epistemic planning they use can model many highly complex scenarios that other types of planning cannot. Likewise, approaches like Argenta and Doyle's (2017) have considerable potential for handling many different TPR problems since they can use just about any type of planning available. Overall, domain-theory based methods are often much more versatile than plan-library based methods.

However, in general, most planning algorithms are more computationally complex than the simple search algorithms used by plan-library based methods. As evidence, both Shvo et al. (2020) and Shvo, Sohrabi, and McIlraith (2018) admitted that their implementations became increasingly more prone to "timing out" as the complexity of their test scenarios increased (e.g., adding more agents, possible actions, etc.). This suggests that domain-theory methods often do not scale well regarding computation time compared to plan-library based methods.

## CONCLUSION

This review aims to bring more interest to the study and implementation of plan recognition in a team setting. Toward this end, we compared and contrasted recent approaches to TPR. Table 1 gives an overview of our surveyed approaches. Ultimately, this survey has helped us identify a few directions for future work.

**Table 1.** Overview of surveyed TPR approaches

| REFERENCE | LIBRARY/PLANNING | SETTING | SUPPORTS |
|---|---|---|---|
| **Plan-Library Based** | | | |
| Banerjee, Kraemer, and Lyle (2010) | Matrices containing grounded actions | Offline | Full Obs. only |
| Sukthankar and Sycara (2011) | AND/OR trees | Offline | Noisy Obs. |
| Banerjee, Lyle, and Kraemer (2015) | Context-free grammar | Offline | Full Obs. only |
| Zhuo (2019) | Matrices containing ungrounded actions | Offline | Noisy/Missing Obs. |
| **Domain-Theory Based** | | | |
| Argenta and Doyle (2017) | Most types | Online | Missing Obs. |
| Shvo, Sohrabi, and McIlraith (2018) | Classical Temporal, Diverse Temporal | Offline | Noisy/Missing Obs. |
| Shvo et al. (2020) | Epistemic | Offline | Noisy/Missing Obs. |

One direction is the development of more online methods. Unfortunately, the only recently published literature on this subject was by Argenta and Doyle (2017). Likewise, Banerjee, Lyle, and Kraemer (2015) offered insight into how their offline method can be adapted for online use. The scarcity of online methods is surprising since it is more beneficial for an AI system to understand a team's behavior and assist them during their task rather than after.

Another direction would be to observe and model team communication as part of the TPR process (Falcone and Castelfranchi, 1996). There has already been some earlier related work in this area. For instance, Kaminka, Pynadath, and Tambe (2002) present a recognizer that infers a team's progress towards a plan from their communications alone. Some other works have the recognizer infer a team's plan from listening to a planning meeting (Kim, Chacha, and Shah, 2015; Kim et al., 2018). Combining concepts from these works with the techniques seen in this review could produce powerful new approaches to handle more realistic and complex team scenarios.

## ACKNOWLEDGMENT

I thank my research advisor, Dr. Clayton Morrison, and our principal investigator, Dr. Adarsh Pyarelal, for their suggestions and feedback in writing this manuscript. This research was conducted as part of DARPA's Artificial Social Intelligence for Successful Teams (ASIST) program, and was sponsored by the Army Research Office and was accomplished under Grant Number W911NF-20-1-0002. The views and conclusions contained in this document are those of the authors and should not be interpreted as representing the official policies, either expressed or implied, of the Army Research Office or the U.S. Government. The U.S.

Government is authorized to reproduce and distribute reprints for Government purposes notwithstanding any copyright notation herein.